# Time-Series Anomaly Classification for Launch Vehicle Propulsion Systems: Fast Statistical Detectors Enhancing LSTM Accuracy and Data Quality

Sean P. Engelstad,[1] Samuel R. Darr,[2] Matthew Taliaferro[3], and Vinay K. Goyal[4]

*The Aerospace Corporation, El Segundo, California 90245*

**Supporting Go/No-Go decisions prior to launch requires assessing real-time telemetry data against redline limits established during the design qualification phase. Family data from ground testing or previous flights is commonly used to detect initiating failure modes and their timing; however, this approach relies heavily on engineering judgment and is more error-prone for new launch vehicles. To address these limitations, we utilize Long-Term Short-Term Memory (LSTM) networks for supervised classification of time-series anomalies. Although, initial training labels derived from simulated anomaly data may be suboptimal due to variations in anomaly strength, anomaly settling times, and other factors. In this work, we propose a novel statistical detector based on the Mahalanobis distance and forward-backward detection fractions to adjust the supervised training labels. We demonstrate our method on digital twin simulations of a ground-stage propulsion system with 20.8 minutes of operation per trial and $O(10^8)$ training timesteps. The statistical data relabeling improved precision and recall of the LSTM classifier by 7% and 22% respectively.**

## I. Introduction

LAUNCH vehicles undergo extensive testing, analysis, and inspections to ensure high-reliability is achieved for successfully deploying their payloads to the intended orbit. Propulsion systems tend to be one of the most challenging areas in the development of new launch vehicles [1-4]. Anomalies identified during ground tests and operations can guide corrective actions that, when implemented, increase launch reliability. Engineers located in mission operations centers monitor thousands of telemetry parameters in real-time. Sensor data is plotted against ground test results, predictions, redlines or historical launch data to assess the health of the system. Redlines are established using qualification testing or analysis [5-7]. Family data are limited in the early phase of a program and redline limits may not be sufficient to detect subtle or unforeseen anomalies for relatively new launch vehicles. In the authors' previous works [8, 9], a machine learning approach for real-time anomaly and identification was developed to assist engineers in their live assessment of telemetry data. The machine learning model is trained on simulated anomalies and can provide real-time anomaly predictions from telemetry data.

---

[1] MTS Graduate Intern, Structures Department, Member AIAA. (Corresponding author, email: sengelstad312@gatech.edu)
[2] Engineering Specialist, Fluid Mechanics Department, Member AIAA.
[3] Engineering Specialist, Fluid Mechanics Department, Member AIAA.
[4] Principal Director, Vehicle Systems Division, Associate Fellow AIAA.

The state-of-the-art in time-series anomaly classification is discussed by Wu et al. [10], and for time-series anomaly detection in Refs. [11, 12]. Common machine learning or ML strategies include reconstruction, forecasting, and direct classification methods. Hundreds of ML models have been used including LSTMs [13], transformers [14], autoencoders [15], variational autoencoders [16], and models that employ wavelet or Fourier transforms [17]. A new strategy [18] with wavelet transform pre-processing, and an LSTM and Transformer architecture, showed superior results against benchmark detection datasets.

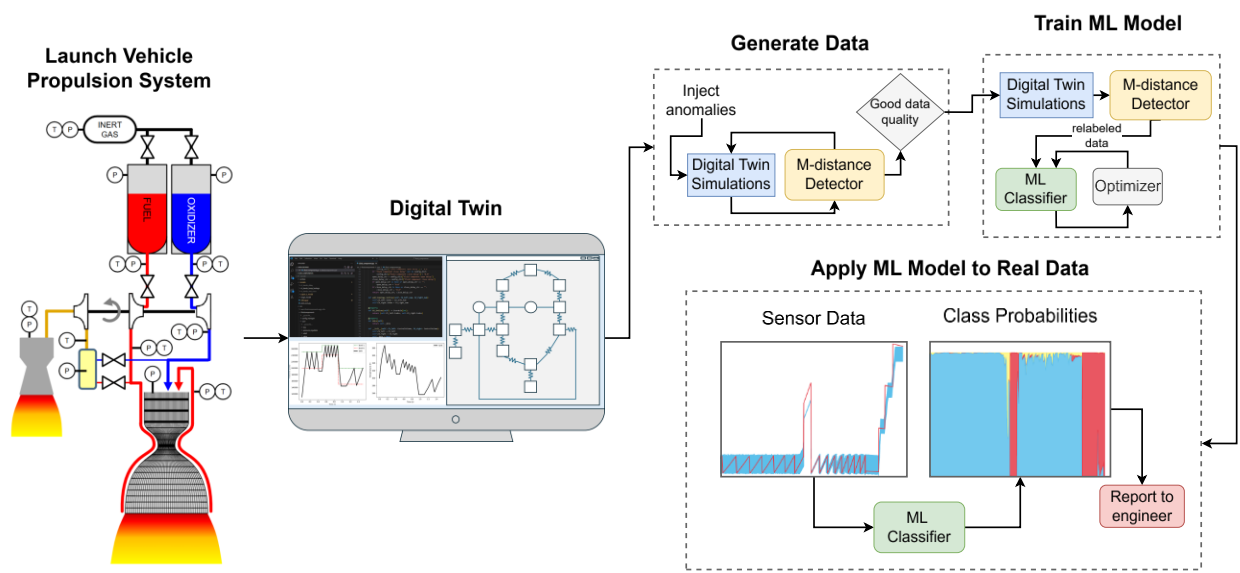

**Fig. 1. Main elements of the paper with simulated anomaly data from a digital twin to train an ML classifier.**

This paper extends the methodology presented in Refs. [8, 9], with a custom (Long Short-Term Memory) LSTM model [13, 26] for supervised anomaly classification of time-series data. The time-series classification problem in this work has over 400 million timesteps, 7 input parameters, and aperiodic valve actuations that generate slope discontinuities in pressure and temperature sensor data. The irregular oscillations result from pressure-band valve logic, which refers to valve cycling that is used to control the pressure of a vessel within a certain target range or band. On smaller datasets from previous work [9], reconstruction and forecasting strategies with transformers and autoencoders were found to struggle with the slope discontinuities from aperiodic valve actuations. Unsupervised methods also tend to have lower precision and may not scale to multiclass detection or classification as well as supervised methods [32]. Also, the continuous wavelet transform [17] increased the dataset size drastically, running into memory limits for this large problem. Thus, the use the custom LSTM model of previous work [9], which achieved a 0.967 F1 score on a small-scale propulsion system dataset, is warranted.

The purpose of this work is to demonstrate a simulated anomaly and machine learning approach on a more realistic propulsion system with more timesteps and complexity in the data. Compared to the previous work [9], the number of input parameters increases from 4 to 7, the number of anomaly classes from 21 to 25, and the number of timesteps from 100 to 2720. In addition, three discrete launch phases are modeled. Anomalies with variable start time and a variety of short, medium and long time scale durations are also considered.

In contrast to the work outlined in [9], the classification labels are corrected with a time-series based anomaly detector. With the large increase from 20 to 101 anomaly generation settings from the author's previous test case, the anomalies were not always detectable depending on the launch phase, valve commands, anomaly strength, or duration. This led to poorer quality of the supervised classification data. Instead of switching to an unsupervised classifier which performs poorly on valve switching data [9], the supervised classification labels are corrected using an unsupervised time-series anomaly detector.

A new strategy [18] with wavelet transform pre-processing, and an LSTM and Transformer architecture, showed superior results against benchmark detection datasets. A lightweight statistics-based detector, using the Mahalanobis distance (M-distance), is preferred over other methods to help adjust the anomaly generation settings over several iterations of dataset generation and quality improvement. Previous M-distance detection methods rely on machine learning models [19, 20, 21] or frequency extractions [22] to obtain anomaly predictions at a given timestep. Time windows and a forward-backwards strategy convert time-covariant M-distances to anomaly predictions at each timestep. This allows us to have accurate detection predictions at each timestep.

The main elements of this work are detailed in Figure 1 which includes a digital twin model of the launch vehicle propulsion system, used to create simulated anomaly data. An unsupervised detector is used to tune anomaly generation settings and relabel the supervised classification data for better data quality. Finally, the LSTM supervised classifier model is trained and applied to real sensor data.

## II. Methodology

### A. Simulation of Fluid System Anomalies

The ground stage propulsion system of Figure 3 is simulated using GiNA [23], a one-dimensional transient CFD solver for launch vehicle propulsion systems of 10 to 100 control volumes. GiNA uses the finite volume method to solve the coupled mass-energy equations. Correlations based on the Rayleigh number are used to predict heat and mass transfer in each ullage, avoiding spatial discretization in typical CFD analyses. An adaptive timestep, with the second order implicit Runge-Kutta scheme, is used to numerically solve the governing equations. GiNA's runtime is driven by property evaluations, not the linear solver, due to the low $O(10^2)$ degree of freedom. GiNA simulations often take on the order of 10 minutes to simulate 2 hours of real time, which allows a large dataset to be generated for machine learning training.

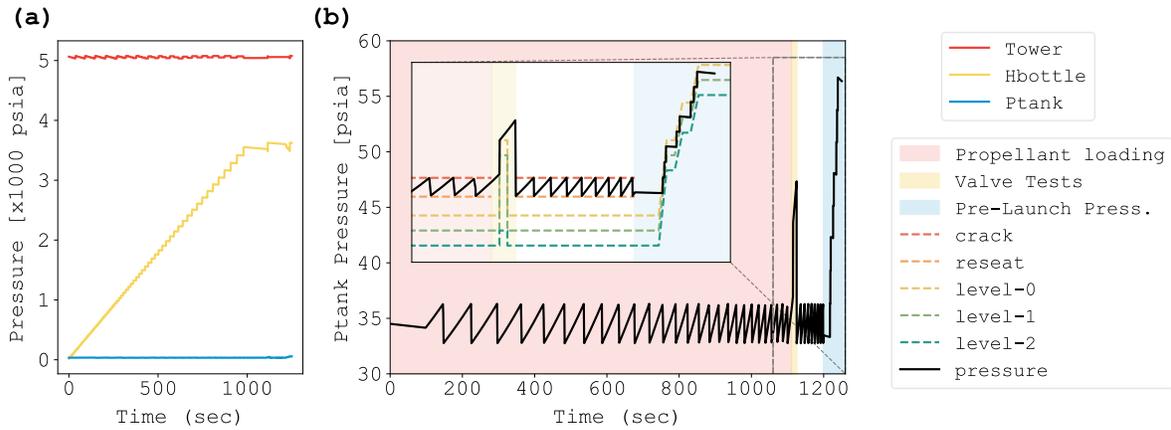

**Fig. 2.** Nominal response of the system with (a) pressures in each control volume, (b) prop tank pressure and valve pressure levels during each launch pressurization phase.

GiNA is augmented with a fluid components (FCs) module, which includes 4 main fluid component types including pressure regulators, shutoff or "bang-bang" valves, variable area valves, and relief valves. Figure 2 shows the fluid components used in the propulsion system for this work – including four shutoff valves SOV-TOW, SOV-HB, SOV-PT1, and SOV-PT2, as well as a lockup relief valve LRV. The pressure-based shutoff valves have a valve state of 0 or 1 determined by pressure band logic. Pressure-based command logic allows a valve to control the tank absolute pressure or a tank-to-tank pressure difference within a preset range or band by cycling open and closed. A lockup relief valve cycles open and closed, keeping the tank pressure between the crack and reseat pressures, unless a lockup command is given which forces the valve to remain closed.

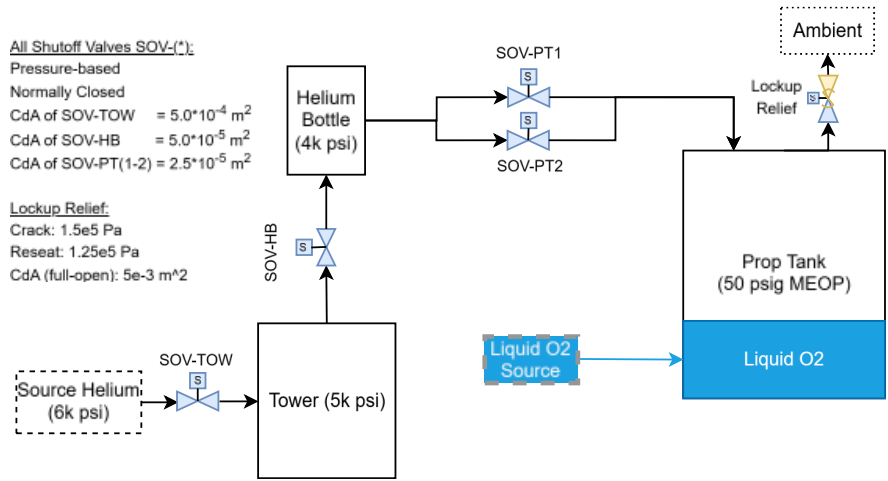

**Fig. 3.** Ground stage propulsion system, with gaseous Helium and Liquid O2 to pressurize a prop tank.

The test case in Figure 3 is intended to emulate a simplified ground-to-vehicle pressurization system. Helium gas is provided by the Source Helium boundary to pressurize the Tower, Helium Bottle, and Propellant Tank. The shutoff valves are used to control the helium flow to their respective downstream volumes. A Lockup Relief Valve (LRV) is used to vent the tank ullage to maintain the pressure between a predefined crack and reseat pressure. The simulation lasts 1250 seconds, with three operation phases shown by the shaded regions of the left hand figure in Figure 2 – propellant loading, valve tests and pre-launch pressurization.

During the propellant loading phase from 0 to 1100 seconds, the Helium Bottle is pressurized and liquid oxygen ($LO_2$) is loaded into the Prop Tank. As the $LO_2$ enters the Prop Tank, the ullage volume decreases and pressure increases, with excess pressure vented through the LRV to ambient. Under nominal operation, the LRV crack and reseats about 50 times during propellant loading, with the time gap between crack and reseats decreasing as the ullage volume contracts.

After 1100 seconds, a valve test is performed. LRV is locked up and SOV-PT1 and SOV-PT2 are opened to pressurize the Prop Tank. At 1125 seconds, LRV is unlocked, and the Prop Tank is depressurized back to the nominal crack-reseat range of LRV until 1200 seconds of the next launch phase.

At 1200 seconds, the pre-launch pressurization phase begins. The Prop Tank is again pressurized by locking up LRV and opening the shutoff valves with several pressure band steps until 1250 seconds. The full nominal simulation takes 1.52 mins runtime for 20.8 minutes of real time.

**B. Anomaly Data Generation**

A large dataset of simulated anomaly data is generated using the Anomaly Data Generation (ADG) module in our code. A Monte Carlo simulation is performed with random anomalies selected from a library of individual component anomalies. A wide range of random inputs for each trial and failure mode are included to make the data more realistic. For example, the initial pressure and temperatures of each control volume are randomized, as are anomaly settings such as leak area, slow opening valve delay, start time and anomaly duration (see Table A.1 in the Appendix for some examples). For many of the settings, a bounded normal distribution at $3\sigma$ is used. The Monte Carlo anomaly simulations used 60 CPU procs to finish 15000 trials in 11.4 hours. Seven input parameters of pressure and temperatures sensors were saved from each trial, including anomaly times for supervised training.

The available anomalies of the propulsion system from Figure 3 are shown in Figure A.1 of the Appendix, with a total of 1 nominal and 24 anomaly classes. There are eight short duration anomalies which last on the order of one second, including fail open and closed failures. The internal leakages are medium duration anomalies of about 40-100 seconds. The remaining anomalies such as slow closing or opening and high crack, etc. last the entire simulation. Although some of the long duration anomalies may only be apparent during the brief times when a valve is actuated. For example, a slow closing anomaly will only materialize when a valve is cycled closed, which may only be a handful of times during the entire simulation.

The 101 anomaly generation settings, with examples in Table A.1, were selected initially from engineering expertise. Then, the M-distance detector, discussed in the next section, was used to identify un-detectable anomalies and adjust anomaly generation settings. The 'un-detectable' anomalies were not identifiable using statistics or human observation. For example, about 40% of the initial class 3 and class 8 internal leakages were not detectable initially. Thus, a leakage study shown in Figure 4 was completed to identify the minimum detectable leakages for each leakage failure and adjust the anomaly generation settings for better data quality. In particular, the leakage rate bounds were then set as a function of the anomaly start time (see Table A.1, Appendix).

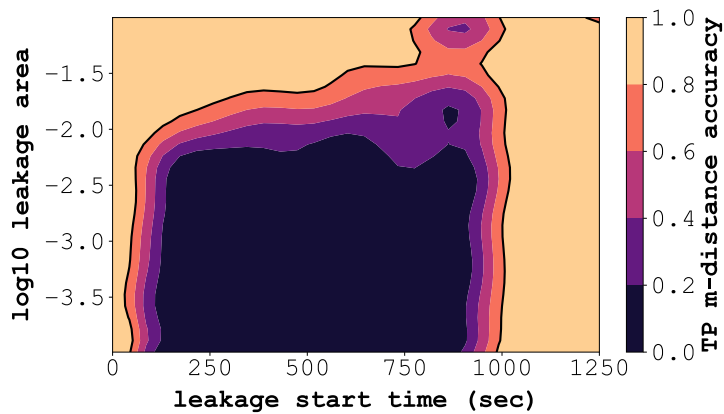

**Fig. 4. Internal leakage study for class 8 the SOV-HB valve.**

## C. Mahalanobis Distance Anomaly Detection

There were several examples in the baseline simulated anomaly dataset where the anomaly was too weak to detect (i.e. indistinguishable from nominal dispersions), or the anomaly took a different amount of time to return to nominal than in our baseline truth labels. To improve data quality, an unsupervised time-series anomaly detector adjusts the anomaly generation settings and relabels the supervised classifier data.

The M-distance measures the normalized distance from the mean in a multivariate normal distribution [28, 29], much like a z-score in a univariate normal distribution. Consider $T$ timesteps in the interval $t \in [a, b]$ such that $x, \mu \in R^T$ are the input state and mean for a single parameter, with $\Sigma \in R^{T \times T}$ the covariance matrix. The mean $\mu$ and covariance $\Sigma$ are computed as sample metrics from the nominal trials in our dataset. The M-distance on a time window $[a, b]$ is then:

$$d_{[a,b]} = \sqrt{(x - \mu)^T (\Sigma + \theta^2 I)^{-1} (x - \mu)|_{t \in [a,b]}} \tag{1}$$

where $\theta = 10^{-4}$ for our dataset and note that the M-distance scales by $1/\theta$ for $x$ outside the covariance eigenbasis.

The dataset has seven input parameters per timestep. As an approximation, each parameter is assumed independent and the max M-distance of each parameter $p$, $d_{[a,b]} = \max_{p} \{d_{[a,b]}^p\}$ is taken. This reduces the cost of inverting the covariance matrix, from size $7T \times 7T$ to size $T \times T$. As the covariance matrix spans across time, contextual anomalies or changes in the curve shape [31] can be picked up, not just point anomalies.

As the data has a high degree of nonlinearity and slope discontinuities, the critical M-distances for a given probability level $P(d < d^*) = P^*$ are not Gaussian. Indeed, the critical M-distance was often 30% or 40% above the Gaussian values. Thus, an adaptive M-distance cutoff $d_{[a,b]}^{nom}(P^*)$ was used from the nominal ensemble data, with probability level $P^* = 0.99$. For convenience, a detection fraction $F_{[a,b]}$ is defined,

$$F_{[a,b]} = \frac{d_{[a,b]}}{d_{[a,b]}^{nom}(P^*)}; \quad ([a,b] \text{ is anom}) \leftrightarrow (F_{[a,b]} > 1) \tag{2}$$

so that values above 1 are anomalies and below 1 are nominal.

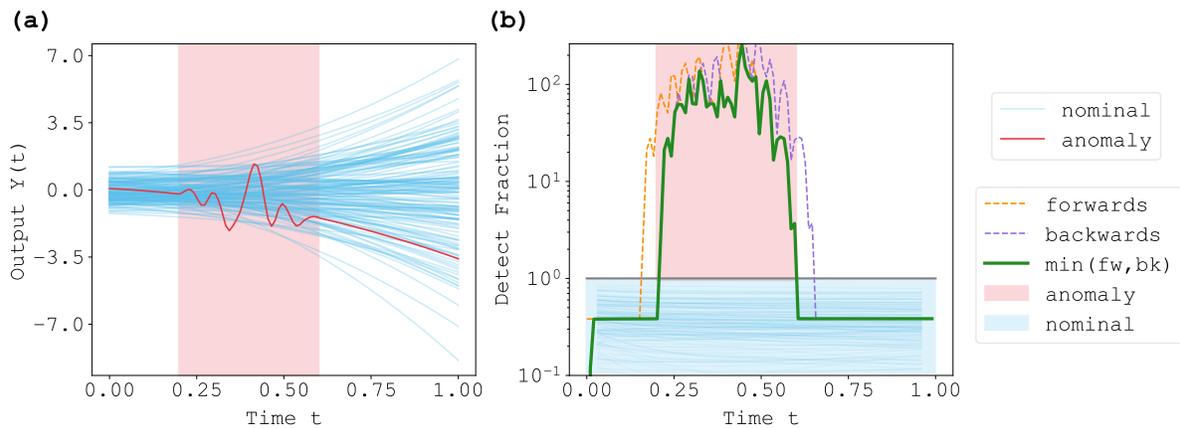

**Fig. 5. M-distance method with (a) nominal and anomaly data from a single output zero mean Gaussian process from Ref. [26, 27] and shown in Table A.2, and (b) detection fractions of the anomaly data.**

However, the detection fraction simply indicates whether the time window $[a, b]$ has an anomaly, not which timestep it occurred. Forwards in time windows $(t, t + T\Delta t]$, can generate false positives before the actual anomaly event. Similarly, backwards in time windows $[t - T\Delta t, t)$ can generate false positives after the anomaly event. To reduce false positives, the minimum detection fraction among forwards and backwards time windows are used,

$$F_{FB}^{(T)}(t) = \min\{F_{[t-T\Delta t, t]}, F_{[t, t+T\Delta t]}\} \tag{3}$$

as in Figure 5. Note how the forward and backward time windows do not agree on the start and end of the anomaly event and the minimum possible anomaly time span is used. This gives a more accurate prediction of the beginning and end of an anomaly event.

The data for Figures 5 and Table A.2 in the Appendix were generated from Gaussian Process kernel functions as described in Refs. [26, 27]. Special time window kernel functions were chosen (see Table A.2 in the Appendix), so that the precise anomaly start and end times were known. Figure A.2 also shows that the shorter detection window sizes $T$ are better as they improve anomaly start and stop time predictions. In the propulsion system dataset of Figure 2, 20 timestep windows are used for short duration anomalies, 100 timestep windows for long duration anomalies and 1 timestep windows for point anomalies. The overall detection fraction is then,

$$F_{\text{ovr}}(t) = \max\left\{F_{FB}^{(1)}(t), F_{\text{FB}}^{(20)}(t), F_{\text{FB}}^{(100)}(t)\right\} \quad (4)$$

which gives conservative anomaly predictions, while considering a variety of time scales. The M-distance method was able to identify each of the 24 anomaly types using the overall detection fractions in Eqn. (4). To improve detector runtime, a stride of 5 was used for the 100 timestep windows, so each covariance matrix was only size $20 \times 20$. Also, multithreading across all 15000 trials improved runtime.

**D. Machine Learning Classification Tool**

The full description of the machine learning classifier model is given in [9]. A custom LSTM [13, 30] model is used with an outer EinsumDense layer in Tensorflow [24]. Time windows of 100 timesteps are used to reduce the number of training weights in the LSTMs. The model inputs are a 3-tensor of normalized input states $\hat{X} \sim (N_b, N_t, N_p)$ with $N_b = 15000$ the number of trials, $N_t = 2721$ the number of timesteps, and $N_p = 7$ the number of input parameters. The model output is a 3-tensor of classification probabilities with $Y \sim (N_b, N_t, N_c)$ and $N_c = 25$ the number of classes. The final dataset of 15000 trials had $2.85 \times 10^8$ training timesteps.

The tensorflow LSTM model was trained with the Categorical Cross Entropy loss function and evaluated by computing the multiclass F1 score [25], with equations given in [9]. Also, in Ref. [9], our LSTM model was shown to be robust to hyperparameters such as dropout, number of layers, learning rate and time window size. The same optimal hyperparameters from the previous study are used as the high training cost makes it infeasible to do hyperparameter tuning. Namely, the model has 3 LSTM layers with [100, 75, 50] neurons each, a learning rate of $3 \times 10^{-4}$, and 100 timestep time windows staggered so only the last 50 timesteps are used for predictions.

## III. Results

**A. Training the Machine Learning Model**

The LSTM Tensorflow model was trained on the full ground stage system dataset with 15000 trials and 25 classes of data, using 750 training epochs in each case and an 80%-10%-10% training-validation-test split. The training was done on two NVIDIA Titan RTX TU102 GPUs, each with 16.31 Tflops and float-32 precision. Training results and metrics are shown in Table I. The training was repeated for both the un-relabeled dataset and the M-distance relabeled dataset, with each ML model training taking about 2.5 days to complete. The M-distance relabeling was done by adjusting the truth class probability labels for each trial to match the detector. The M-distance relabeling for the $O(10^8)$ timesteps took only 1.486 hours to complete thanks to speedup from multithreading, jit python compilation and covariance strides. Table I shows that the train and test loss are significantly reduced from relabeling, with a 12% improvement in test F1 score as well.

Table 1 - Training metrics for the LSTM model

| Dataset | Relabeling time, hrs | Training Time, hrs | Train (Test) Loss | Test F1 Score |
| --- | --- | --- | --- | --- |
| Un-relabeled | n/a | 62.849 | 0.512 (0.449) | 0.710 |
| Relabeled | 1.486 | 61.075 | 0.165 (0.326) | 0.831 |

M-distance relabeling adjusts the supervised classifier truth labels to match the detector's anomaly vs nominal predictions. Figure 6 shows an example where the baseline truth labels for class 2 and class 3 failures are not accurate to the actual anomaly durations. For example, with the class 2 fail open anomaly, the valve was failed open for only 1.3 seconds (a very short duration) but the high system residuals last the rest of the simulation. Even though the valve is no longer failed, the high system residuals are outside the nominal distribution and are labeled an anomaly by the M-distance detector. While the class 2 and class 3 failures are distinguishable during the valve failure, they are very similar after the failure with slowly decreasing press-tower and temp-tower residuals. Thus, for anomaly trials with similar high system residual signatures, the F1 scores of the LSTM model were manually adjusted to allow similar class predictions after the valve failure. Often times, the fail open, internal leakage and slow opening anomalies on the same component had similar post-failure residual signatures.

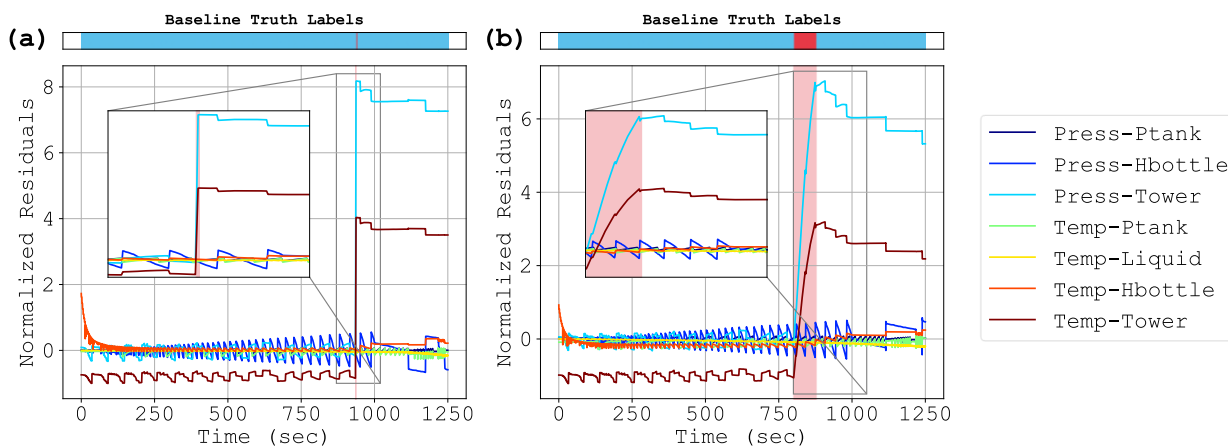

**Fig. 6. Residual states for (a) class 2 fail open and (b) class 3 internal leakage. The anomaly signatures differ during the valve failure but appear similar after the failure with high residuals.**

Overall metrics on the test data are shown for the 'no-relabel', 'relabel' and 'relabel + corrected' datasets in Table II. Here, 'relabel' indicates that the M-distance relabeling was performed, while 'corrected' indicates the manual corrections for post-failure class predictions. The results show that individual class accuracies significantly improve by relabeling and the post-failure corrections. The precision improves by 7%, indicating fewer false positives, while the recall improves by 22%, indicating fewer false negatives. Thus, it appears the greatest effect of relabeling is to remove false negatives by extending anomalies. Also, the test F1 score improves 9% by increasing the dataset size for the relabeled + corrected case.

It is hypothesized that the F1 score increased with additional trials, as 7500 trials may not fully span the space of anomalies - considering the 101 anomaly generation settings and 3 launch phases in our data.

Table 2 - Test data metrics of the LSTM Model By dataset

| Number of Trials | Data Type | Precision | Recall | F1 |
|---|---|---|---|---|
| 7500 | Un-relabeled | 0.741 | 0.522 | 0.613 |
| | Relabeled | 0.815 | 0.733 | 0.772 |
| | Relabeled+ corrected | 0.828 | 0.759 | 0.792 |
| 15000 | Un-relabeled | 0.850 | 0.665 | 0.703 |
| | Relabeled | 0.879 | 0.820 | 0.833 |
| | Relabeled+ corrected | **0.916** | **0.858** | **0.876** |

The final confusion matrix after the M-distance relabeling and post-failure corrections is shown in Figure 7 for the test dataset of 1500 trials (10% of the data). The corresponding F1 score for the confusion matrix is 88%, with a precision of 92% and recall of 86%. The high precision indicates very few false positives. The lower recall indicates more false negatives occur than false positives. It is likely that the remaining false negatives of column 0 could be significantly reduced by tuning more of the anomaly generation settings with M-distance and increasing the number of trials.

B. Example Applications to Test Data

Next, consider three examples to show the effect of M-distance relabeling on class probability predictions: (1) Figure 8 shows a trial which has high system residuals that persist after the valve failure; (2) Figure 9 shows a permanently failed valve which only produces an effect on the system during valve actuation times; and (3) Figure 10 shows another trial with high post-failure system residuals, which required a manual correction to the classification errors from Figures 6 and Table II.

First, in Figure 8, a class 21 slow-opening anomaly of LRV is shown. The valve fails for 1.47 seconds, resulting in an increase in Prop Tank pressure that takes 31 seconds to return to nominal. The M-distance method accurately predicts the time for the pressure to return to nominal. Also, the classification signature of the LSTM model has fewer false positives and false negatives for the relabeled than the non-relabeled case, making the anomaly easier to identify for a human observer.

In Figure 9, a class 14 slow opening anomaly of SOV-PT 1 and 2 is shown. The failure mode is applied to the valves for the entire simulation; however, the failure only manifests in off-nominal behavior during the valve actuation times of the valve test and pre-launch pressurization launch phases. The M-distance predictions for off-nominal behavior match when the anomaly is outside the nominal family data. In this example, the anomaly is correctly predicted in the non-relabeled (Figure 9c) and relabeled cases (Figure 9d). However, the relabeled ML model (Figure 9d) has fewer false positives, and would be more reliable for the engineer. Also, if relabeling was not performed, some training timesteps that are within the nominal distribution would be incorrectly labeled as anomalous, thus contaminating the dataset and resulting in poorer ML predictions.

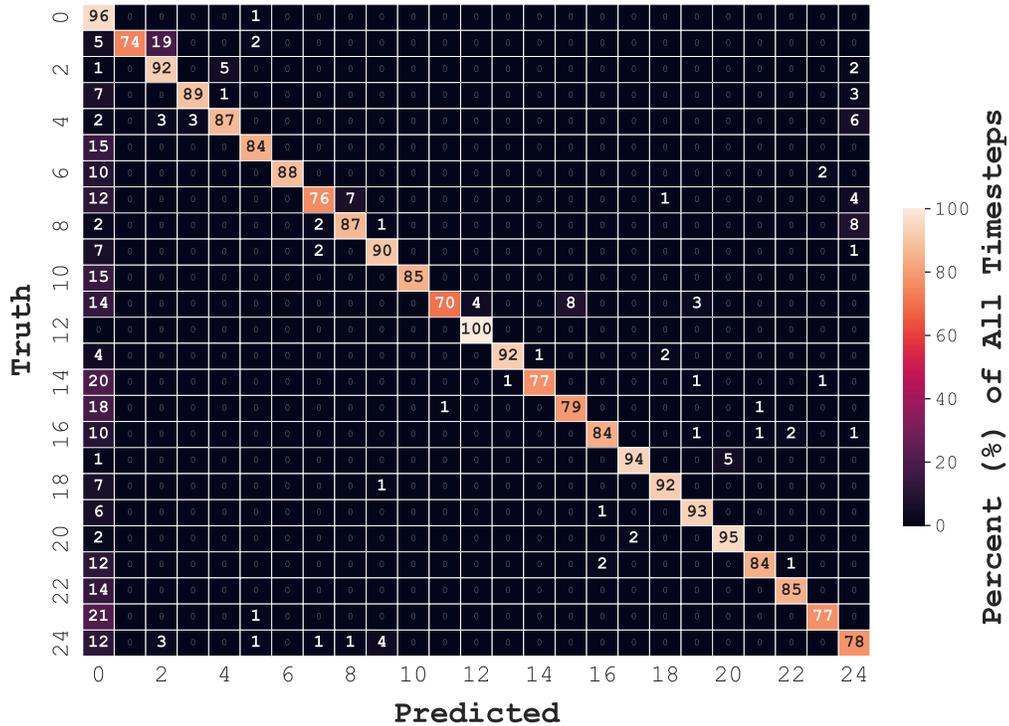

**Fig. 7. Final confusion matrix after M-distance relabeling and manual post-failure corrections.**

In Figure 10, a class 2 fail open anomaly SOV-TOW is shown. The valve is failed for only 1.23 seconds, but this results in a high Tower pressure that doesn't return to nominal the rest of the simulation. As shown in Figure 6, the class 2 and class 3 failures have similar anomaly signatures after the valve is failed and with high system residuals. Thus, a manual correction to the classifier error was performed on this trial to accept class 3 predictions past the valve failure. In each of the three trials of Figures 8-10, the M-distance relabeling accurately predicted the off-nominal behavior and reduced false positive and false negative rates of the LSTM model. This shows the importance of improving data quality for developing practical automatic anomaly identification tools for launch vehicle propulsion systems.

## IV. Conclusion

The combined ML and physics-based tool developed in our previous work [8, 9] was applied to the anomaly classification of a

Approved for public release. OTR 2025-00908

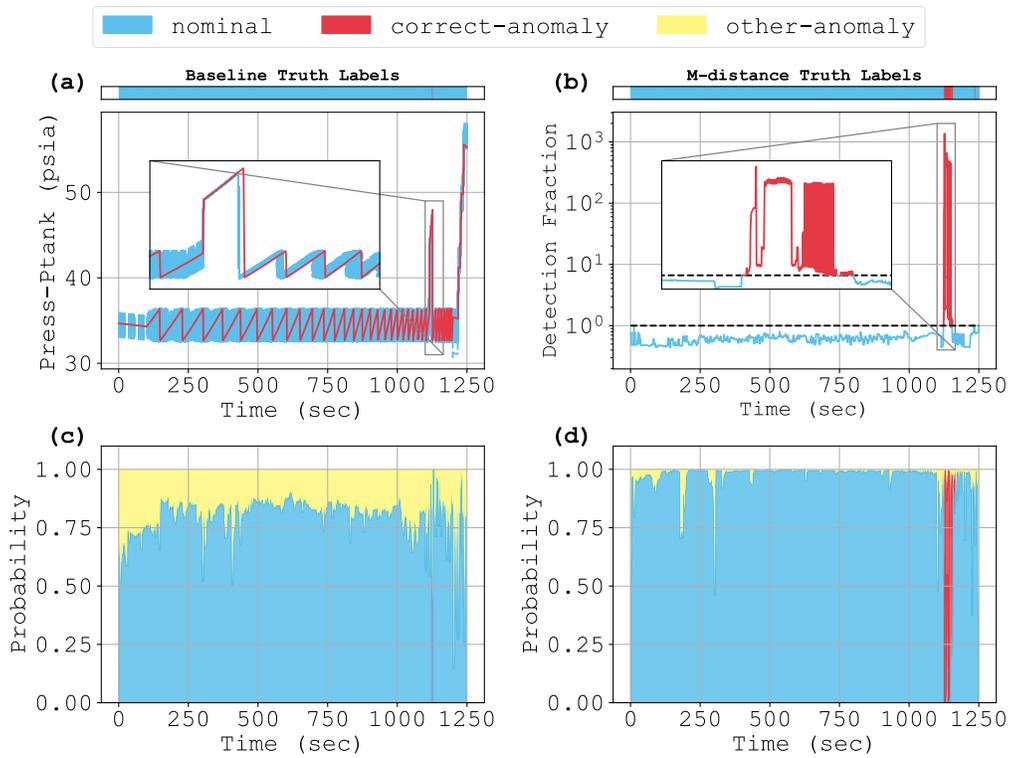

**Fig. 8.** Slow-opening anomaly of the LRV valve with (a) raw sensor data, (b) M-distance detection, (c) and (d) ML predictions before and after relabeling.

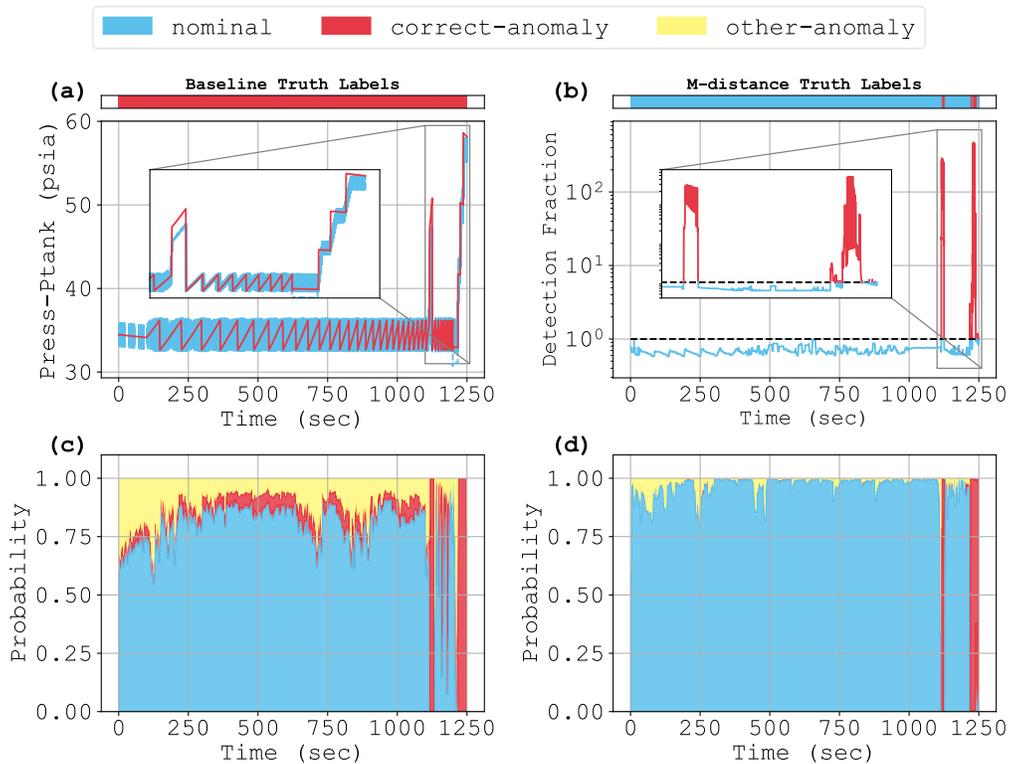

**Fig. 9.** Slow-opening of the SOV-PT1,2 valves with (a) raw sensor data, (b) M-distance detection, (c) and (d) ML predictions before and after relabeling.

large-scale ground-to-vehicle pressurization system, with more realistic pressure band valves and longer simulation times than the previous test case [9]. The training dataset consisted of over 400 million time steps, 24 failure modes, 20.8 minutes of real-time simulation per trial, and a wide range of short and long time-scale simulated anomalies. 101 anomaly generation settings were included to produce a wide range of strength, duration and start times for each simulated anomaly. After it was found that the ML model performed poorly on the baseline labels, an M-distance unsupervised time-series detector method was developed to correct the supervised training labels. The M-distance method was also used to improve anomaly generation settings by finding the minimum detectable leakage areas. The results showed the M-distance method improved the test F1 scores of our LSTM models by 12% with no adjustment for acceptable class confusions, and 16% with the adjustment. In several example trials, the M-distance method correctly identified in which pre-launch phases the anomaly was detectable, could extend anomalies with high system residuals, and could find the minimum detectable anomaly strength.

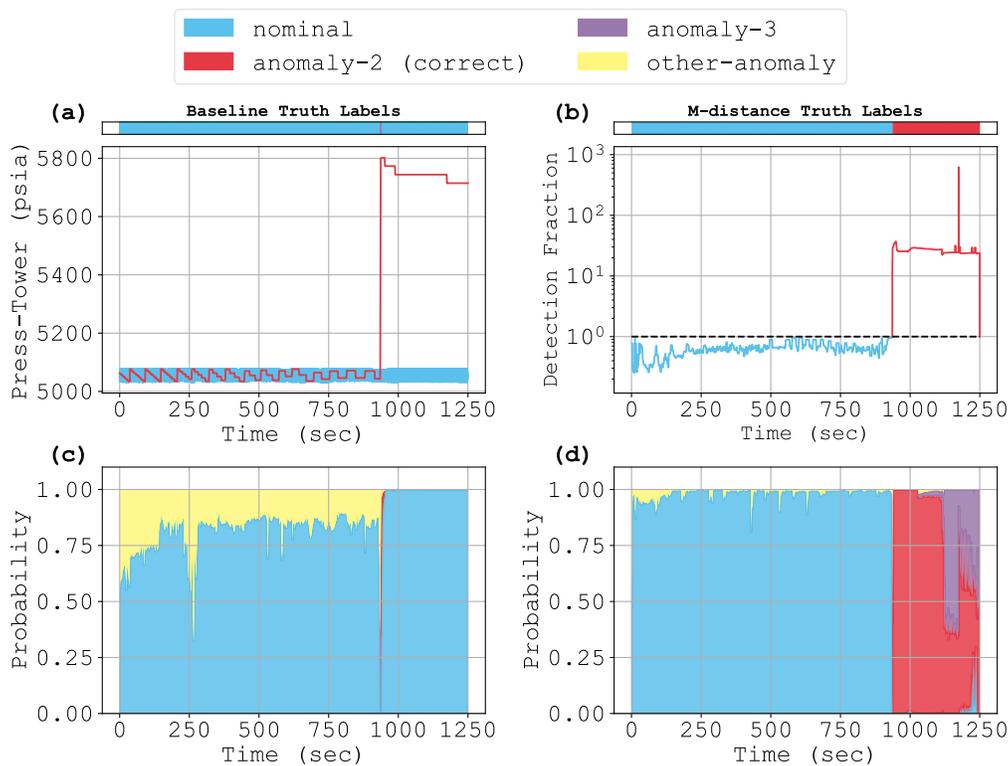

**Fig. 10. Fail open anomaly of the SH-TOW valve with (a) raw input data, (b) M-distance detection, (c) and (d) ML predictions before and after relabeling.**

The M-distance method significantly improved the quality of our supervised training data, decreasing the false positive rate of our LSTM models. Also, with the final LSTM model trained on the relabeled data and manually corrected, the F1 score is 88%, demonstrating an effective classifier model. This shows the importance of improving data quality to developing practical automatic anomaly identification tools for launch vehicle propulsion systems.

A. **Practical Lessons Learned**

The following are practical lessons learned from our work:

1. The M-distance method should be used to identify minimum detectable leakage area as a function of time. The pressure difference across the leak area is subject to change and may be too small during certain phases to cause a detectable leak rate if the leak area is kept constant. This ensures the appropriate detectable ranges are included for training the ML model.

2. Dataset size is very important to obtain good M-distance and ML model training results. Three hundred trials per anomaly was found to be a good starting size for a dataset, although six hundred trials per anomaly were needed for better accuracy in this dataset. M-distance false positives may indicate under-sampling and the need for more trials.

3. If the nominal variation of an input parameter is too low, the M-distance method can report weak anomalies after the valve is no longer in a failed state that would not be considered an anomaly during launch. These weak anomalies hurt the ML model performance after relabeling. Thus, initial pressure and temperatures of each control volume and crack and reseat pressures were chosen with sufficiently large standard deviations to avoid such errors. It is helpful to plot the nominal ensemble data and check if the plot region is densely covered enough.

4. The M-distance method can correct anomaly class labels by extending anomalies that produce a persistent off-nominal signature after the valve is no longer failed. However, after the valve is no longer in a failed state, several anomalies that produce high residuals look identical and are confused by the ML model, and indeed they are not distinguishable even by a human. For example, the fail open, internal leakage and slow closing anomalies on the same component have identical signatures after the valve failure. In practice, the human monitor can rely on the first anomaly class that is reported to be the most probable, since that should correspond to the time when the anomaly is active. Human expertise can also provide judgement on which anomalies produces similar post-anomaly signatures to help discern the true cause of the failure.

# Appendix

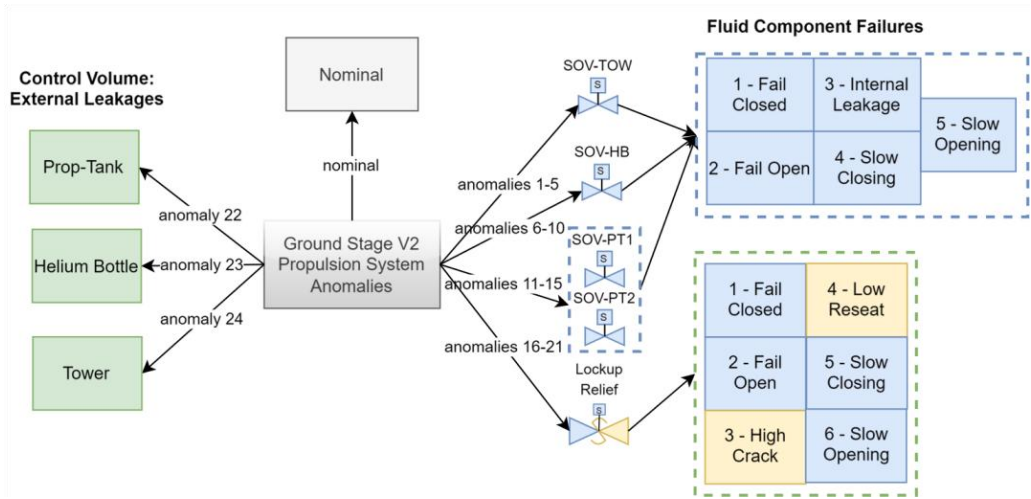

**Fig. A.1. Available anomalies for the ground stage propulsion system in Figure 3.**

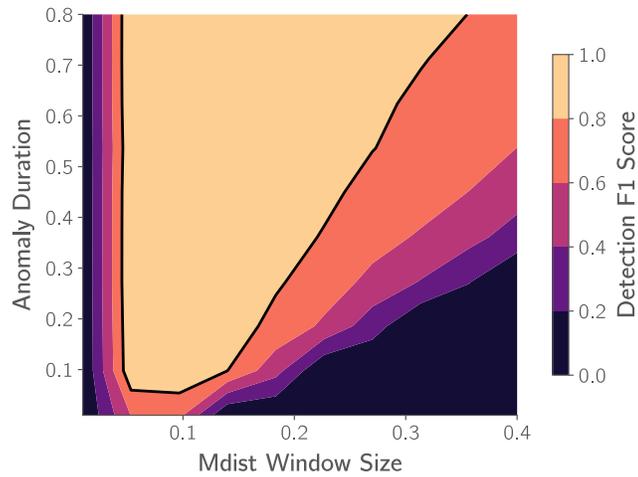

**Fig. A.2. M-distance F1 scores as a function of detection window size and anomaly duration for GP kernel data.**

**Table A.1 - Random anomaly generation settings, with $H(t)$ the heaviside function for t>1000 sec**

|  | SOV-TOW | SOV-HB | SOV-PT1,2 | LRV |
|---|---|---|---|---|
| Fail Open Fraction | [0.1, 0.3] | [0.1, 0.3] | [0.1, 0.3] | [0.1, 0.5] |
| Slow opening delay (s) | [0.1, 0.25] | [0.1, 0.25] | [0.025, 0.1] | [1.0, 3.0] |
| Slow closing delay (s) | [5, 10] | [1.0, 2.5] | [1.5, 3.0] | [5.5, 7.5] |
| Log10 Internal Leak Fraction | $\begin{bmatrix}-2.8 - 1.2H(t),\\ -1.8 - 1.2H(t)\end{bmatrix}$ | $\begin{bmatrix}-2.3 - .7H(t)\\ -1.7 - .3H(t)\end{bmatrix}$ | [-2.52, -2.0] | --- |

**Table A.2 – Single outputs from zero mean Gaussian processes, e.g. $Y_{nom}(t) \sim N(0, k_{nom}(t, t'))$, using time window kernels from Ref. [26, 27].**

| Kernel | Expression |
| --- | --- |
| $k_{nom}(t, t')$ | $0.3 + 0.4tt' + 3(tt')^2 + t(t')^2 + t'(t)^2$ |
| $W_{[a,b]}(t, t')$ | $relu\left(1 - \left|\frac{2t - (a+b)}{b-a}\right|\right) \cdot relu\left(1 - \left|\frac{2t' - (a+b)}{b-a}\right|\right)$ |
| $k_{anom}(t, t')$ | $k_{nom}(t, t') + 0.3 \cdot W_{[a,b]}(t, t') \cdot \exp[-2 \cdot sin^2(10\pi|t - t'|)]$ |

## Acknowledgments

This work was performed under an Internal Research and Development funding at The Aerospace Corporation. No other funding sources were used to advance this technology.